\documentclass[runningheads]{llncs}
\usepackage{graphicx}
\usepackage{amsmath,amssymb} 
\usepackage{epstopdf}
\usepackage{color}

\usepackage{cite}
\usepackage{subfigure}
\usepackage{multirow}
\usepackage{diagbox}
\usepackage{rotating}
\usepackage{booktabs}
\usepackage{colortbl}
\usepackage{overpic}
\usepackage[table]{xcolor}
\usepackage{textcomp}
\usepackage{contour}
\usepackage{textcomp}

\definecolor{mygreen}{RGB}{0,150,0}
\definecolor{myred}{RGB}{200,0,0}
\usepackage[pagebackref=false,letterpaper=true,colorlinks=true,linkcolor=myred,citecolor=blue,bookmarks=false]{hyperref}

\renewcommand{\tabcolsep}{.5mm}

\newcommand{\eg}{\emph{e.g.}}
\newcommand{\ie}{\emph{i.e.}}
\newcommand{\etal}{\emph{et al.}}

\graphicspath{{./Imgs/}}

\newcommand\blfootnote[1]{%
\begingroup
\renewcommand\thefootnote{}\footnote{#1}%
\addtocounter{footnote}{-1}%
\endgroup
}

\begin{document}
\mainmatter

\title{Cross-Modal Weighting Network for \\ RGB-D Salient Object Detection} 
\titlerunning{CMWNet for RGB-D SOD} 
\author{Gongyang Li\inst{1} \and
Zhi Liu\inst{1,*} \and
Linwei Ye\inst{2} \and \\
Yang Wang\inst{2,4} \and
Haibin Ling\inst{3}}
%
\authorrunning{G. Li et al.}

\institute{Shanghai University, Shanghai, China \and
University of Manitoba, Winnipeg, Canada \and 
Stony Brook University, NY, USA \and 
Huawei Technologies Canada\\
\email{ligongyang@shu.edu.cn, liuzhi@staff.shu.edu.cn}\\
\email{\{yel3,ywang\}@cs.umanitoba.ca, hling@cs.stonybrook.edu}\\
\url{https://github.com/MathLee/CMWNet}}

\maketitle              
\begin{abstract}
Depth maps contain geometric clues for assisting Salient Object Detection (SOD).
In this paper, we propose a novel Cross-Modal Weighting (CMW) strategy to encourage comprehensive interactions between RGB and depth channels for RGB-D SOD.
Specifically, three RGB-depth interaction modules, named CMW-L, CMW-M and CMW-H, are developed to deal with respectively low-, middle- and high-level cross-modal information fusion.
These modules use Depth-to-RGB Weighing (DW) and RGB-to-RGB Weighting (RW) to allow rich cross-modal and cross-scale interactions among feature layers generated by different network blocks.
To effectively train the proposed Cross-Modal Weighting Network (CMWNet), we design a composite loss function that summarizes the errors between intermediate predictions and ground truth over different scales.
With all these novel components working together, CMWNet effectively fuses information from RGB and depth channels, and meanwhile explores object localization and details across scales.
Thorough evaluations demonstrate CMWNet consistently outperforms 15 state-of-the-art RGB-D SOD methods on seven popular benchmarks.
\blfootnote{* Zhi Liu is the corresponding author.}

\end{abstract}

\section{Introduction}
Salient object detection (SOD) aims to pick the regions/objects in an image that are most attractive
to human visual attention.
It has a wide range of applications as summarized in recent
surveys~\cite{2015SODBenchmark,Borji2019,2019sodsurvey}.
Most existing SOD solutions take as input an RGB image (or video), which is convenient in many
application scenarios, but may suffer from challenges such as low contrast and disturbing background.
Alternatively, one can seek help from the depth information typically provided with an RGB-D input.
In fact, with the popularity of depth sensors/devices, RGB-D SOD has received extensive attention recently,
and numerous
approaches~\cite{Niu2012STEREO,Li2014LFSD,Feng2016LBE,Song2017MDSF,Qu2017DF,Han2018CTMF,Chen2018PCF,Zhao2019CPFP,Cong2019DTM,Fan2019D3Net}
have been proposed to extract salient objects from paired RGB images and depth maps.

Starting with the first stereoscopic image SOD dataset STEREO~\cite{Niu2012STEREO},
traditional RGB-D SOD methods mainly apply contrast
cue~\cite{Fang2014TIP,Fan2014DSP,SHK2015Saliency},
fusion framework~\cite{Peng2014NLPR,Guo2015Salient,Song2017MDSF,2017MBD} and
measure strategy~\cite{Li2014LFSD,Cong2016SPL,Feng2016LBE} to
extract the complementary information in depth maps.
These well-designed hand-crafted features-based methods,
which are influenced by RGB SOD solutions, have achieved remarkable results.
However, salient objects in the generated saliency maps are
sometimes blocky because of inappropriate over-segmentation, while salient objects may be confused by complex scenes.

Recently, with the rapid development of deep learning,
convolutional neural networks (CNNs) have shown strong dominance
in many computer vision problems.
Many CNN-based RGB-D SOD methods have been proposed and greatly outperformed traditional ones.
Early CNN-based methods~\cite{Qu2017DF,Shigematsu2017ICCVW}
feed the superpixel-based hand-crafted features of RGB-D pairs into CNNs,
but their results are still patch-based.
Subsequent methods instead assign saliency values for each pixel based
on the RGB image and depth map in an end-to-end manner.
Among these methods, the two-stream
architecture~\cite{Han2018CTMF,Chen2018PCF,Wang2019AFNet,Chen2019MMCI,Ding2019JVCIR,2019PDNet}
fuses cross-modal features/saliency maps in the middle/late stage,
while the single-stream architecture~\cite{Liu2019SRCNN,Fan2019D3Net}
directly handles RGB-D pairs.
These methods, despite achieving great performance gain, do not take full advantage of rich interactive information between different modalities and scales of CNN blocks.

Motivated by the above observation, in this paper, we propose a novel
\emph{Cross-Modal Weighting Network} (CMWNet)
that significantly improves RGB-depth interactions, and hence boosts RGB-D SOD performances as demonstrated in our thorough experiments.
Our key idea is to jointly explore the information carried by both RGB and depth channels, and to encourage cross-modal and cross-scale RGB-depth interactions among different CNN feature blocks. This way, our algorithm can capture both microscopic details carried by shallow blocks and macroscopic object location information carried by deep blocks.
CMWNet adopts a three-level representation, capturing low-, middle-, and high-level information respectively; and multiple blocks at different scales are allowed to be within a level. The cross-modal cross-scale interactions are modeled through the novel Cross-Modal Weighting (CMW) modules to highlight salient objects.

In particular, we propose three CMW modules, CMW-L, CMW-M and CMW-H.
For low- and middle-level parts, CMW-L and CMW-M are used to enhance salient object details in a cross-scale manner.
For high-level part, CMW-H is used to enhance salient object localization,
which plays a crucial role in subsequent prediction of salient objects.
The key components in these CMW modules are the proposed Depth-to-RGB Weighting (DW) and RGB-to-RGB Weighting (RW) operations that enhance RGB features in each channel  based on corresponding response maps.
In addition to the encoder, a three-level decoder is designed to connect the
three-level enhanced features to predict the final salient objects.
In this way, the proposed CMWNet effectively exploits the properties
of CNN features and strengthens the cross-modal and
cross-scale interactions, resulting in excellent performance.

Our major contributions are summarized as follows:
\begin{itemize}
	\setlength{\itemsep}{0pt} \setlength{\parsep}{1pt} 	\setlength{\parskip}{1pt}
	\item We explore the complex complementarity between RGB image and depth map
	 in a three-level encoder-decoder structure, and propose a novel \emph{Cross-Modal Weighting Network} (CMWNet) to encourage the cross-modal and cross-scale interactions, boosting the performance of RGB-D SOD.
	\item We propose three novel RGB-depth interaction modules to effectively enhance both salient object details (CMW-L and CMW-M) and salient object localization (CMW-H). 
	\item Extensive experiments on seven popular public datasets under six commonly used evaluation metrics show that the proposed method achieves the best performance compared with 15 state-of-the-art RGB-D SOD methods.
\end{itemize}


\section{Related Work}

\noindent\textbf{Traditional RGB-D SOD.}
Starting from the first work for saliency detection~\cite{1998Itti},
the contrast-based approaches are the mainstream for saliency detection.
This trend has spread to traditional RGB-D SOD.
Numerous contrast-based RGB-D SOD methods have been proposed,
such as disparity contrast~\cite{Niu2012STEREO},
depth contrast~\cite{Fang2014TIP,Cheng2014DES,Fan2014DSP,SHK2015Saliency},
and multi-contextual contrast~\cite{Peng2014NLPR}.
Song \etal~\cite{Song2017MDSF} employed the multi-scale fusion
to merge saliency maps to obtain the final RGB-D saliency map, which is
similar to methods based on two-stream saliency fusion~\cite{Peng2014NLPR}
and multiple-cues fusion~\cite{Guo2015Salient,2017MBD}.
By adopting the objectness measure~\cite{Li2014LFSD},
depth confidence measure~\cite{Cong2016SPL} and
salient structure measure~\cite{Feng2016LBE}, the performance gets clear improvement.
Besides, other methods (\ie,
global prior~\cite{Ren2015CVPRW}, cellular automata~\cite{Guo2016ICME},
transformation strategy~\cite{Cong2019DTM})
have been proposed for RGB-D SOD.
However, these traditional methods are often based on superpixels,
regions and patches, which cause saliency maps to appear blocky
and saliency values to be scattered.
%

\noindent\textbf{CNN-based RGB-D SOD.}
In recent years, numerous CNN-based RGB-D SOD methods~\cite{Qu2017DF,Shigematsu2017ICCVW,Han2018CTMF,Chen2018PCF,2019PDNet,Chen2019MMCI,Ding2019JVCIR,Chen2019TANet,Zhao2019CPFP,Wang2019AFNet,Liu2019SRCNN,Fan2019D3Net}
have been proposed.
As pioneering work based on CNNs, Qu \etal~\cite{Qu2017DF}
fed the superpixel-based RGB-D saliency features into a
five-layer CNN.
Shigematsu \etal~\cite{Shigematsu2017ICCVW} sent
ten superpixel-based depth features to a network.
Being patch-based methods, these methods sometimes generate results that appear blocky.
To overcome the limitation, Han \etal~\cite{Han2018CTMF} proposed a transfer and fusion
based network to predict pixel-level saliency values.
The single-stream architecture~\cite{Liu2019SRCNN,Fan2019D3Net}
adopts a straightforward way to handle the four-channel RGB-D pair.
This architecture does not effectively capture the
cross-modal interactions between the RGB image and
the depth map, so the performance depends largely
on the network structure rather than the cross-modal interactions.
The two-stream architecture employs two separate networks
to extract features~\cite{Han2018CTMF,Chen2018PCF,2019PDNet,Chen2019MMCI}
and saliency maps~\cite{Ding2019JVCIR,Wang2019AFNet},
and then fuse them with various strategies.
Some works~\cite{Ding2019JVCIR,Wang2019AFNet,Han2018CTMF,2019PDNet}
only fuse saliency maps and high-level features.
As a result, they do not capture more complex cross-modal
interactions at other levels of the network.
Some other works~\cite{Chen2018PCF,Chen2019TANet} consider
cross-modal CNN features, but the same module is used to
process cross-modal CNN features at different blocks.
Consequently, these methods ignore the different properties
of CNN features at different blocks and cannot provide specific enhancements to
object details and object localization.

In this work, we propose a novel three-level CMWNet to encourage interactions between RGB and depth channels and propose several modules to treat differently detail features and localization features carried in CNN feature blocks at various scales.
Moreover, we process CNN features in a cross-modal and cross-scale manner
to effectively capture the interactions across modalities and scales.
Thus, our network can accurately enhance the details and localization of salient objects in the
encoder part and precisely infer salient objects in the three-level decoder.

\section{Proposed Method}
\label{sec:OurMethod}

In this section, we start with the overview of Cross-Modal
Weighting Network (CMWNet) (\S~\ref{sec:Overview}).
In \S~\ref{sec:CCW}, we provide the details of low- and middle-level cross-modal
weighting modules, \ie~CMW-L and CMW-M, and then in \S~\ref{sec:CW} we introduce the high-level cross-modal
weighting module CMW-H.
Finally, we describe the implementation details in \S~\ref{sec:Details}.
%

\begin{figure*}[t!]
	\centering
	\begin{overpic}[width=\textwidth]{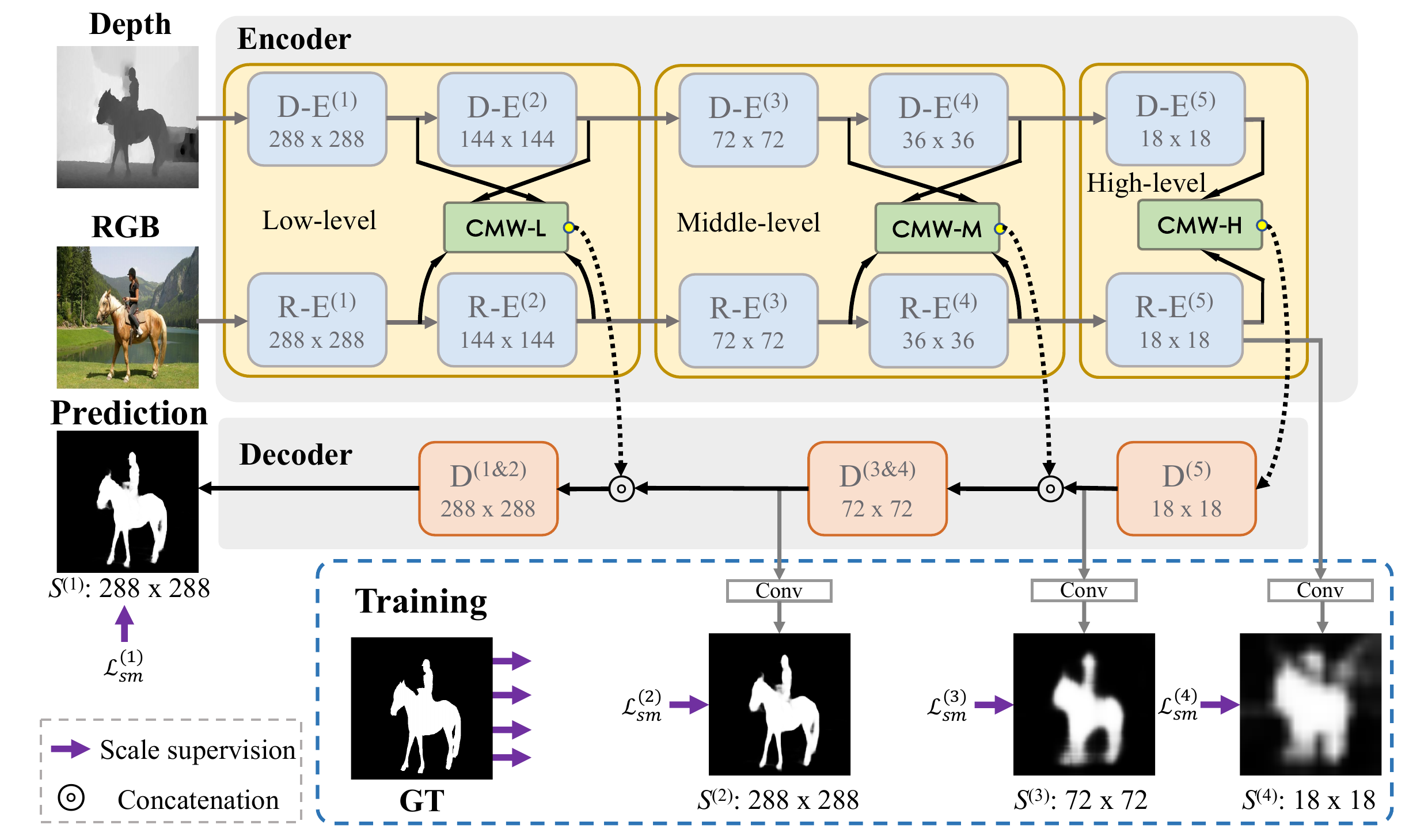}
    \end{overpic}
	\caption{\small \textbf{Illustration of the proposed CMWNet.}	
	For both RGB and depth channel, the Siamese encoder network is employed to extract feature blocks organized in three levels.
	Three Cross-Modal Weighting (CMW) modules, CMW-L, CMW-M and CMW-H, are proposed to capture the interactions at corresponding level, and provide inputs for the decoder. The decoder progressively aggregates all the cross-modal cross-scale information for the final prediction.
	For training, multi-scale pixel-level supervision for intermediate predictions are utilized.
    }
    \label{fig:Framework}
\end{figure*}

\subsection{Network Overview and Motivation}
\label{sec:Overview}

The proposed CMWNet follows a three-level Siamese encoder-decoder structure, as summarized in Fig.~\ref{fig:Framework}.

\noindent\textbf{Three-level Encoder.}
We adopt the VGG16~\cite{2014VGG16ICLR} as the backbone.
The depth map branch and the RGB image branch share the same weights.
In the Siamese encoder part,
five CNN blocks of the depth map and the RGB image are
denoted as D-E$^{(l)}$ and R-E$^{(l)}$ ($\mathit{l}\in$\{1, 2, 3, 4, 5\}
is the block index), respectively.
Considering the unique properties of features, we divide the first
and second CNN blocks into low-level part, the third and fourth CNN blocks
into middle-level part, and the last CNN block into high-level part.

\noindent{\textbf{Low- and Middle-level Cross-Modal Weighting Modules.}}
The weighting mechanism~\cite{2018Weighting} is an extended version
of attention mechanism, and it aims to modulate features of each channel
according to particular response maps.
The abundant geometric knowledge of depth maps is helpful to provide
object details and object localization for SOD.
We novelly extend the weighting mechanism with cross-modal information
(\ie~RGB image and depth map), and propose cross-modal RGB-depth interaction modules,
which adopt weighting mechanism to reweight the RGB features
based on depth response maps and RGB response maps to focus
on salient objects.

Considering that the low- and middle-level parts carry abundant information about object details, we treat the features in these two levels as responsible for object details enhancement, and propose CMW-L and CMW-M.
Each of the low- and middle-level contains the two adjacent CNN blocks, one contains relatively macroscopic information while the other relatively microscopic.
To balance these two types of information within a level, we use the higher \textit{depth} block to enhance the lower \textit{RGB} block, and use the lower \textit{depth} block to enhance the higher \textit{RGB} block, namely \textit{cross-scale Depth-to-RGB weighting}.
It is an important component of CMW-L and CMW-M.
Concretely, the higher depth response maps are in charge of modulating the lower RGB features,
and the lower depth response maps are responsible to modulate the higher RGB features.
Such a cross-scale way captures cross-scale complementarity of cross-modal features.
Besides, the Depth-to-RGB weighting is executed between two adjacent blocks, which can
capture the continuity of features.

On the other hand, RGB features have the ability to modulate themselves.
For this purpose, we introduce the \textit{RGB-to-RGB weighting} to CMW-L and CMW-M.
RGB features are enhanced by RGB response maps, which are generated from RGB features.
This allows our weighting modules to learn and adjust salient parts in an adaptive manner.
Depth-to-RGB weighting and RGB-to-RGB weighting complement each other, improving the stability and robustness of our inference.
Thus, for example, in CMW-M,
R-E$^{(3)}$ is enhanced by D-E$^{(4)}$ and R-E$^{(3)}$, and
R-E$^{(4)}$ is enhanced by D-E$^{(3)}$ and R-E$^{(4)}$.
Notably, CMW-L and CMW-M perform the same cross-scale scheme, but with different resolutions.
The multi-resolution enhanced object details of features benefit SOD.

\noindent\textbf{High-level Cross-Modal Weighting Module.}
The high-level part is distinct from the other two parts,
and it contains rich global information.
Thus, in the high-level part, we adopt CNN features of the highest blocks
(\ie, D-E$^{(5)}$ and R-E$^{(5)}$)
to accurately locate salient objects.
We propose the CMW-H module, which is the modified variant of CMW-L and CMW-M,
to enhance the macroscopic localization of salient objects.
The RGB features of R-E$^{(5)}$ are enhanced by the depth response maps generated from
D-E$^{(5)}$ and the RGB response maps generated from R-E$^{(5)}$.

\noindent\textbf{Three-level Decoder.}
The decoder can make good use of features from the encoder
with skip-connections.
To fuse all the enhanced features for effective inference,
the decoder part also consists of three levels,
\ie, D$^{(5)}$, D$^{(3\&4)}$ and D$^{(1\&2)}$ as shown in Fig.~\ref{fig:Framework},
corresponding to high-level, middle-level and low-level encoder parts.
Between the two adjacent levels, we employ the deconvolutional layer
for 4$\times$ upsampling.
Specifically, to effectively train the proposed CMWNet, we adopt the deep scale
supervision~\cite{2015DeepSup} behind each level to force features of the decoder
network to focus on salient objects.

\begin{figure*}[t!]
	\centering
	\begin{overpic}[width=\textwidth]{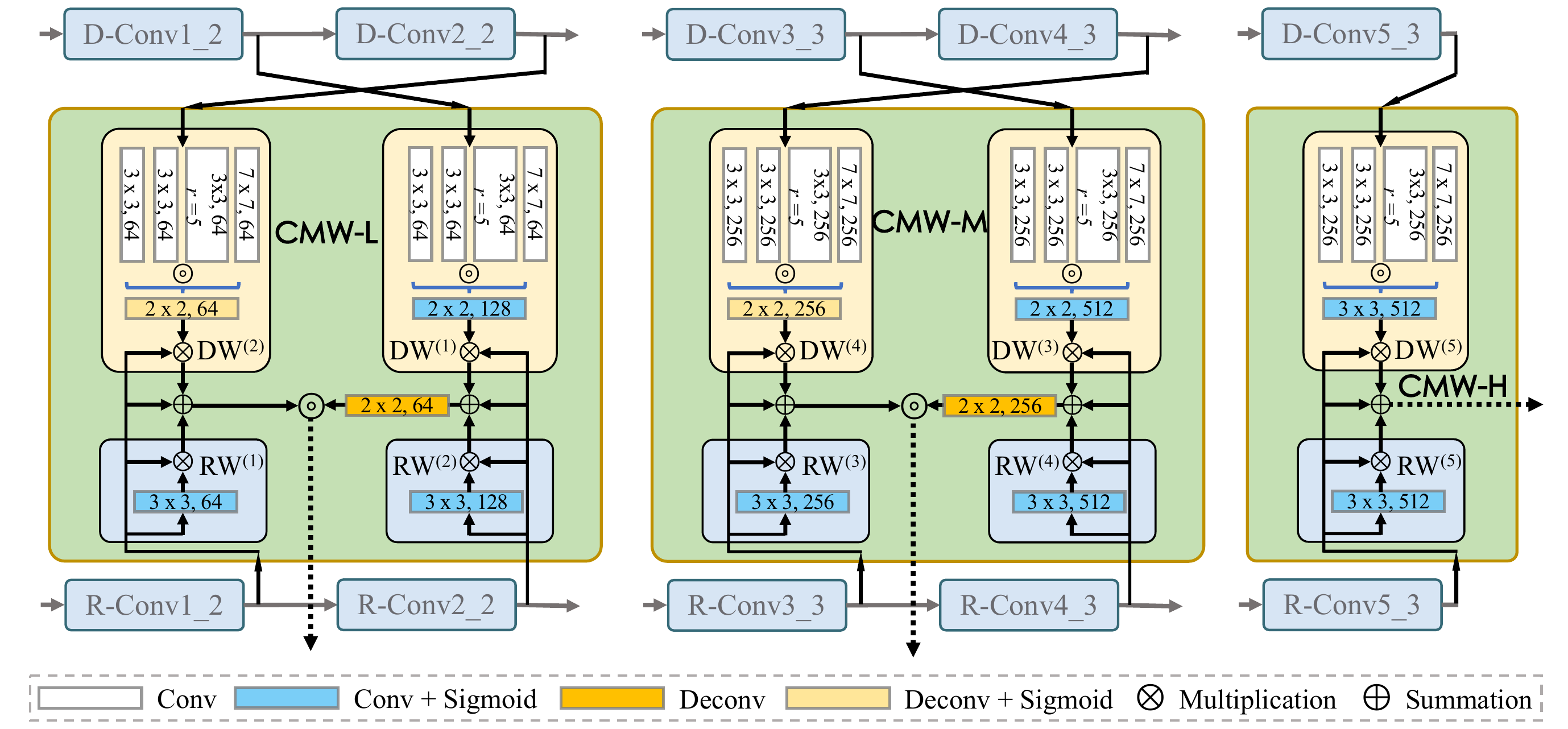}
    \end{overpic}
	\caption{\small \textbf{Details of the three proposed RGB-depth interaction modules: CMW-L, CMW-M and CMW-H.}	
    All modules consist of Depth-to-RGB Weighting (DW) and RGB-to-RGB Weighting (RW) as key operations.
	Notably, the DW in CMW-L and CMW-M is performed in the cross-scale manner between two adjacent blocks, which effectively captures the feature continuity and activates cross-modal cross-scale interactions.
    }
    \label{fig:CMW}
\end{figure*}

\subsection{Low- and Middle-level Cross-Modal Weighting}
\label{sec:CCW}

We perform the enhancement on RGB features at low-level and
middle-level parts with the CMW-L and CMW-M, respectively.
The details of \textbf{CMW-L} and \textbf{CMW-M} are shown
in Fig.~\ref{fig:CMW}.
There are two types of weighting in each module, \ie,
cross-scale Depth-to-RGB weighting (DW) and RGB-to-RGB weighting (RW).
For each encoder block, those two types of weighting
only apply to the last feature layer.
So we denote the last layer of features in D-E$^{(l)}$ and R-E$^{(l)}$
as $\mathbf{f}^{(l)}_{d}$ and $\mathbf{f}^{(l)}_{r}$, respectively.

We provide a simplified version to explain the principle of
DW and RW.
For a feature map $\mathbf{F}\!\in\!\mathbb{R}^{C_0\!\times\!W_0\!\times\! H_0}$,
there are two groups of  weighting response maps
$\mathbf{r}_1\!\in\![0,1]^{C_0\!\times\!W_0\!\times\! H_0}$ and
$\mathbf{r}_2\!\in\![0,1]^{C_0\!\times\!W_0\!\times\! H_0}$
to modulate it at pixel level, and then the two types of weighting can be formulated as:
%
\begin{equation}
   \begin{aligned}
    \mathbf{EF} = \mathbf{F} + \mathbf{r}_1\otimes\mathbf{F} + \mathbf{r}_2\otimes\mathbf{F},
    \label{eq:principle}
    \end{aligned}
\end{equation}
where $\mathbf{EF}\!\in\!\mathbb{R}^{C_0\!\times\!W_0\!\times\! H_0}$
is the enhanced feature, $\otimes$ is the element-wise multiplication,
and + is the element-wise summation.
$\mathbf{r}_1\otimes\mathbf{F}$ and
$\mathbf{r}_2\otimes\mathbf{F}$ can be regarded as the
DW operation and RW operation, respectively.
If $\mathbf{r}_1$ and $\mathbf{r}_2$ have good responses to
salient objects (\ie, the pixel value is close to 1 on salient objects
and close to 0 on background), $\mathbf{F}$ will be accurately
modulated to focus on the desired salient parts and $\mathbf{EF}$
will have a stronger representation for salient objects.
Thus, we apply the two types of weighting to RGB features
and depth features for enhancement of salient object details in the CMW-L and CMW-M
modules.

\noindent\textbf{Depth-to-RGB Weighting.}
In our network, the Depth-to-RGB weighting is the most important
operation to mine the complementarity of depth maps.
It works in a cross-modal and cross-scale manner.
To expand the receptive field and increase the feature diversity,
we design a comprehensive structure of filters to generate
the local and global features $\mathbf{f}^{(l)}_{lg}$.
Concretely, we adopt two convolutional layers
with $3\!\times\!3$ kernel as local filters.
We also adopt a dilated convolution~\cite{Dila2016} with $3\!\times\!3$ kernel and $rate\!=\!5$
and a convolutional layers with $7\!\times\!7$ kernel
as global filters, as shown in DW$^{(l)}$ of Fig.~\ref{fig:CMW}.
The global filters in the comprehensive structure expand
the receptive field of convolution operations.
The obtained global features can capture macro-level information of depth
features, which are complementary to the local features.
For each $\mathbf{f}^{(l)}_{d}$, $\mathbf{f}^{(l)}_{lg}$ can be computed as:
%
\begin{equation}
   \begin{aligned}
    \mathbf{f}^{(l)}_{lg} = \mathrm{concat}\big(  C(\mathbf{f}^{(l)}_{d};\mathbf{W}^{(l_1)}_{loc}),C(\mathbf{f}^{(l)}_{d};\mathbf{W}^{(l_2)}_{loc}),
      C(\mathbf{f}^{(l)}_{d};\mathbf{W}^{(l_1)}_{glo}),C(\mathbf{f}^{(l)}_{d};\mathbf{W}^{(l_2)}_{glo}) \big),
    \label{eq:SD-DW}
    \end{aligned}
\end{equation}
where $\mathrm{concat}(\cdot)$ denotes the cross-channel concatenation,
$C(\ast;\mathbf{W}^{(l_i)}_{loc})$ is a convolutional layer with
parameters $\mathbf{W}^{(l_i)}_{loc}$
(\ie, $\mathbf{W}^{(l_1)}_{loc}$ and $\mathbf{W}^{(l_2)}_{loc}$ are $3\!\times\!3$ kernel)
for producing local features,
$C(\ast;\mathbf{W}^{(l_1)}_{glo})$ is a convolutional layer with
parameters $\mathbf{W}^{(l_1)}_{glo}$
(\ie, $\mathbf{W}^{(l_1)}_{glo}$ is $7\!\times\!7$ kernel),
and $C(\ast;\mathbf{W}^{(l_2)}_{glo})$ is the dilated convolution with
parameters $\mathbf{W}^{(l_2)}_{glo}$
(\ie, $\mathbf{W}^{(l_2)}_{glo}$ is $3\!\times\!3$ kernel with $rate\!=\!5$).

Then, the multi-scale features in $\mathbf{f}^{(l)}_{lg}$ are fused to generate
the depth response maps $\mathbf{r}^{(l)}_{dw}$.
Specifically, to make $\mathbf{r}^{(l)}_{dw}$ have the
same resolution as the corresponding cross-scale RGB features,
the fusion operation is a convolutional layer with stride 2 for 2$\times$ downsampling for DW$^{(1)}$ and DW$^{(3)}$,
while a deconvolutional layer is used for DW$^{(2)}$ and DW$^{(4)}$ for 2$\times$ upsampling.
Thus, $\mathbf{r}^{(l)}_{dw}$ can be computed as:
%
\begin{equation}
   \begin{aligned}
    \mathbf{r}^{(l)}_{dw}=\left\{
	\begin{array}{rcl}
	\sigma(C(\mathbf{f}^{(l)}_{lg};\mathbf{W}^{(l)}_{dw})),      & l=1,3\\
	\sigma(De(\mathbf{f}^{(l)}_{lg};\mathbf{W}^{(l)}_{dw})),      & l=2,4\\
	\end{array} , \right. 
    \label{eq:DRMap}
    \end{aligned}
\end{equation}
where $\sigma(\cdot)$ is the sigmoid function,
and $De(\ast;\mathbf{W}^{(l)}_{dw})$ is the deconvolutional layer with
parameters $\mathbf{W}^{(l)}_{dw}$, which are $2\!\times\!2$ kernel with stride 2.

Finally, $\mathbf{r}^{(l)}_{dw}$ is used to enhance the
cross-scale RGB features as follows:
%
%
\begin{equation}
   \begin{aligned}
    \mathbf{f}^{(l)}_{dw}=\left\{
	\begin{array}{rcl}
	\mathbf{r}^{(l+1)}_{dw}\otimes\mathbf{f}^{(l)}_{r},      & l=1,3\\
	\mathbf{r}^{(l-1)}_{dw}\otimes\mathbf{f}^{(l)}_{r},      & l=2,4\\
	\end{array} . \right. 
    \label{eq:DW}
    \end{aligned}
\end{equation}

\noindent\textbf{RGB-to-RGB Weighting.}
Considering that the RGB features of low- and middle-level parts also contain
rich information about details of salient objects, we
propose the RGB-to-RGB weighting to adaptively enhance RGB features
with the RGB response maps $\mathbf{r}^{(l)}_{rw}$,
which are generated from $\mathbf{f}^{(l)}_{r}$ as follows:
%
\begin{equation}
   \begin{aligned}
    \mathbf{r}^{(l)}_{rw} = \sigma(C(\mathbf{f}^{(l)}_{r};\mathbf{W}^{(l)}_{rw})).
    \label{eq:SRMap}
    \end{aligned}
\end{equation}
The details of filters in RW$^{(l)}$ are also presented in Fig.~\ref{fig:CMW}.
Then, similar as depth response maps,
$\mathbf{r}^{(l)}_{rw}$ can enhance $\mathbf{f}^{(l)}_{r}$ as follows:
%
%
\begin{equation}
   \begin{aligned}
    \mathbf{f}^{(l)}_{rw} = \mathbf{r}^{(l)}_{rw}\otimes\mathbf{f}^{(l)}_{r}.
    \label{eq:SW}
    \end{aligned}
\end{equation}
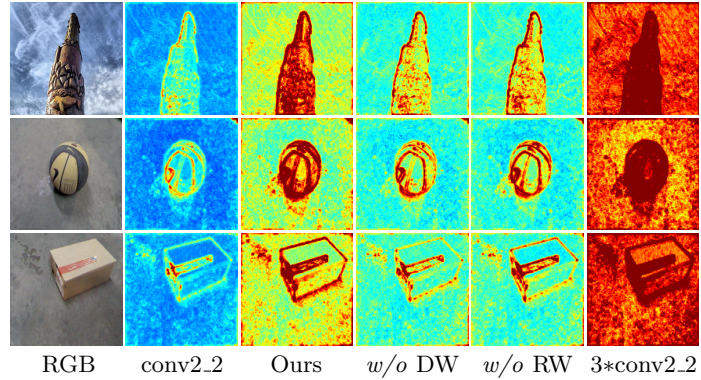
\begin{figure}[t!]
    \centering
    \normalsize
	\begin{overpic}[width=.75\textwidth]{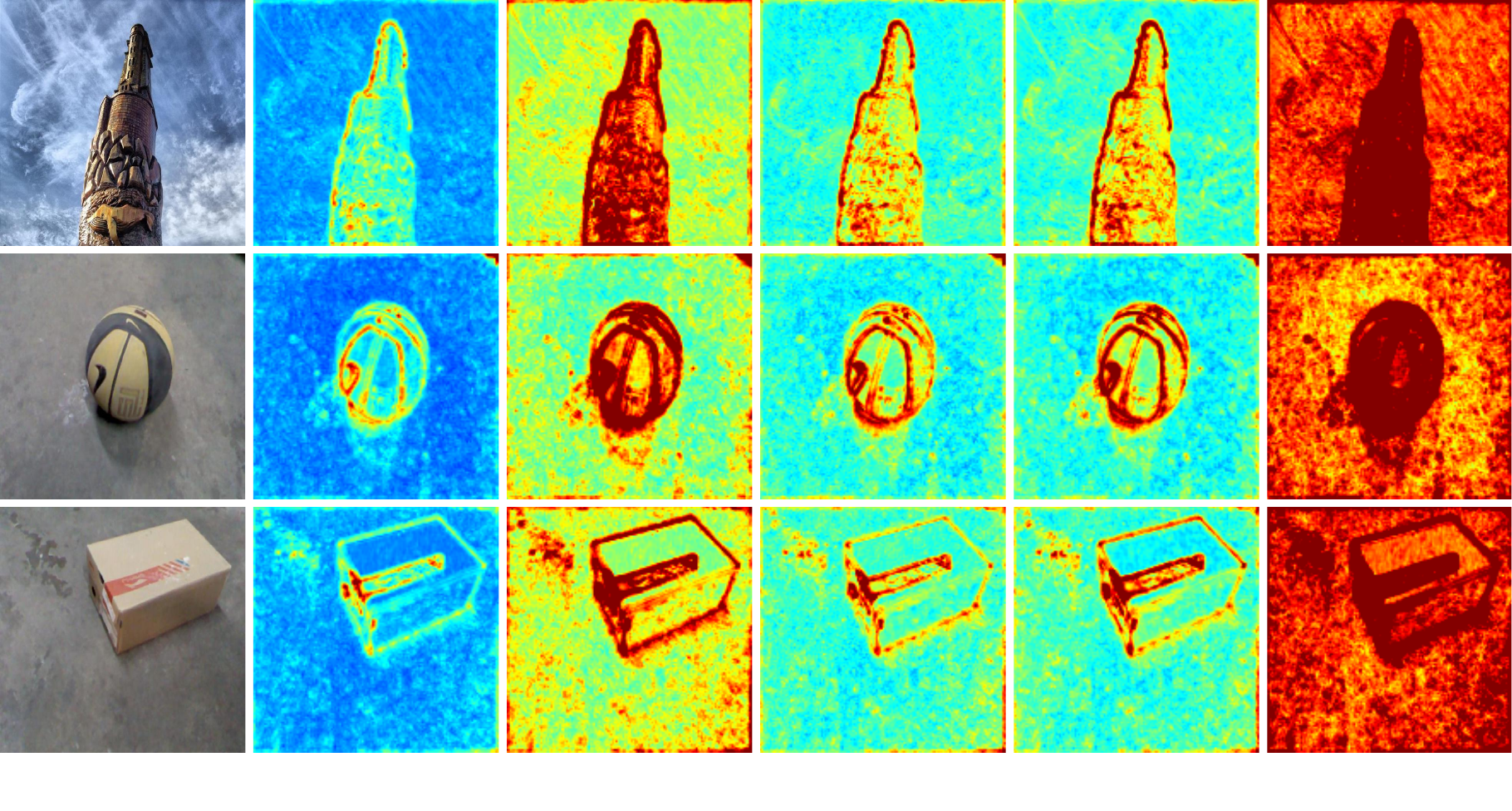}
    \put(4.6,0){\small RGB}
    \put(19,0){\small conv2\_2}
    \put(37.8,0){\small Ours}
    \put(51.6,0){\small \textit{w/o} DW}
    \put(68.5,0){\small \textit{w/o} RW}
    \put(84.3,0){\small 3$\ast$conv2\_2}
    \end{overpic}
	\caption{\small \textbf{Visualizing features of RGB conv2\_2 in CMW-L.}
    \textit{w/o} DW: without adding DW features;
    \textit{w/o} RW: without adding RW features;
    3$\ast$conv2\_2: features with triple linear enhancement.
    }
    \label{fig:FeaturemapVisualize}
\end{figure}

\noindent\textbf{Aggregation of Double Weighting Features.}
After the DW and RW operations,
the RGB features are enhanced twice and can capture
the details of salient objects.
To preserve the original color information, we add
RGB features to these two groups of enhanced features to produce
the double enhanced features $\mathbf{f}^{(l)}_{de}$.
The aggregation of double weighting features is defined as:
%
%
\begin{equation}
   \begin{aligned}
    \mathbf{f}^{(l)}_{de} = \mathbf{f}^{(l)}_{r} + \mathbf{f}^{(l)}_{dw} + \mathbf{f}^{(l)}_{rw}.
    \label{eq:DualE}
    \end{aligned}
\end{equation}
Combining Eq.~\ref{eq:DW}, Eq.~\ref{eq:SW} and Eq.~\ref{eq:DualE},
we find that Eq.~\ref{eq:DualE} is similar to Eq.~\ref{eq:principle}.
Thus, for the CMW-L and CMW-M, the output
features $\mathbf{f}^{(k)}_{cmw}$ can be computed as:
%
%
\begin{equation}
   \begin{aligned}
    \mathbf{f}^{(k)}_{cmw} = Cat( \mathbf{f}^{(2k-1)}_{de},	De(\mathbf{f}^{(2k)}_{de};\mathbf{W}^{(2k)}_{cmw})),  &     &   k=1,2.
    \label{eq:CCW}
    \end{aligned}
\end{equation}
Then, $\mathbf{f}^{(2)}_{cmw}$ and $\mathbf{f}^{(1)}_{cmw}$
boost the salient object inference in D$^{(3\&4)}$ and D$^{(1\&2)}$, respectively,
through skip-connections (\ie, the dashed line in Fig.~\ref{fig:Framework}).
More detailed parameters of CMW-L and CMW-M are shown in Fig.~\ref{fig:CMW}.

In Fig.~\ref{fig:FeaturemapVisualize}, we visualize features of
RGB conv2\_2 in \textbf{CMW-L} to verify the effectiveness of
the double weighting enhancement.
By comparing ``conv2\_2" and ``Ours", salient objects are
highlighted more clearly in ``Ours".
If we delete $\mathbf{f}^{(l)}_{dw}$ (\textit{w/o} DW) or
$\mathbf{f}^{(l)}_{rw}$ (\textit{w/o} RW), salient objects are more indistinct than ``Ours".
We also show the features with triple linear enhancement (``3$\ast$conv2\_2") to
demonstrate that the double weighting enhancement is more effective
than the conventional linear enhancement.

\subsection{High-level Cross-Modal Weighting}
\label{sec:CW}
As for the high-level part, we modify the cross-scale CMW-L to the same-scale 
manner to effectively utilize the macroscopic semantic information.
We propose the CMW-H for object localization enhancement.
For DW operation in CMW-H, the RGB features are enhanced by
depth response maps of the same layer.
So, the DW operation in Eq.~\ref{eq:DW} is modified as follows:
%
%
\begin{equation}
   \begin{aligned}
    \mathbf{f}^{(l)}_{dw} = \sigma(C(\mathbf{f}^{(l)}_{lg};\mathbf{W}^{(l)}_{dw})) \otimes \mathbf{f}^{(l)}_{r},  &     & l=5.
    \label{eq:DRMapCW}
    \end{aligned}
\end{equation}
Other operations, such as the RW and features aggregation,
are the same as those in CMW-L.
Notably, the output of CMW-H is $\mathbf{f}^{(5)}_{de}$,
which is directly fed to the decoder part.
It leads the inference process of SOD,
as shown in Fig.~\ref{fig:Framework}.
The detailed structure of CMW-H is present in Fig.~\ref{fig:CMW}.

\subsection{Implementation Details}
\label{sec:Details}

\noindent\textbf{Loss Function.}
As shown in Fig.~\ref{fig:Framework}, we add a convolutional layer after
R-E$^{(5)}$, D$^{(5)}$ and D$^{(3\&4)}$ to generate
intermediate predictions $S^{(4)}$, $S^{(3)}$ and $S^{(2)}$ at the
training phase.
Then, we utilize different scales of ground truth (GT) to supervise them
and the final prediction $S^{(1)}$ with the softmax loss.
The total loss $\mathbb{L}$ can be defined as:
%
%
\begin{equation}
   \begin{aligned}
    \mathbb{L}  = \sum\limits_{t=1}^4 \alpha_{t} \cdot  \mathcal{L}^{(t)}_{sm} (S^{(t)},G^{(t)}),
    \label{eq:TotalLoss}
    \end{aligned}
\end{equation}
where $\mathcal{L}^{(t)}_{sm} (\cdot,\cdot)$ is the softmax loss,
$\alpha_{t}$ is the loss weight and set to 1,
and $G^{(t)}$ is a GT of the same resolution as $S^{(t)}$.

\noindent\textbf{Network Training Protocol.}
Our CMWNet is implemented in Caffe~\cite{2014Caffe}
with an NVIDIA Titan X GPU.
The parameters of the Siamese encoder part are initialized by
the VGG16 model~\cite{2014VGG16ICLR}, except that the conv1\_1
of the depth stream is initialized by the Gaussian distribution with
a standard deviation of 0.01.
Other newly added layers are initialized using the
Xavier initialization~\cite{2010Xavier}.
Following~\cite{Han2018CTMF,Fan2019D3Net}, the training set
consists of 1,400 triplets from NJU2K~\cite{Ju2014NJU2K}
and 650 triplets from NLPR~\cite{Peng2014NLPR}.
We resize all training triplets to 288$\times$288, and then
we adopt the rotation ($90^{\circ}$, $180^{\circ}$ and $270^{\circ}$)
and mirror reflection for augmentation, resulting in 10.25K training triplets.
We employ the SGD~\cite{2010SGD} to train the network for 22.5K
iterations.
The learning rate, batch size, iteration size, momentum and
weight decay are set to $10^{-7}$, 1, 8, 0.9 and 0.0001, respectively.
The learning rate will be divided by 10 after 12.5K iterations.

\section{Experiments}

\subsection{Datasets and Evaluation Metrics}
\noindent\textbf{Datasets.}
We evaluate the proposed method and all compared methods on seven public benchmark datasets,
including STEREO~\cite{Niu2012STEREO}, NJU2K~\cite{Ju2014NJU2K},
LFSD~\cite{Li2014LFSD}, DES~\cite{Cheng2014DES},
NLPR~\cite{Peng2014NLPR}, SSD~\cite{SSD2017}
and SIP\cite{Fan2019D3Net}.
%

%
%
%
%
%
%
%
%
%

\noindent\textbf{Evaluation Metrics.}
We evaluate the performance of our method and other methods using
six widely used evaluation metrics including
maximum F-measure ($\mathcal{F}_{\beta}$, $\beta^2$ = 0.3),
weighted F-measure ($\mathcal{F}_{\beta}^w$, $\beta^2$ = 1)~\cite{2014WeiFm},
mean absolute error (MAE, $\mathcal{M}$),
precision-recall (PR) curve,
S-measure ($\mathcal{S}_{\lambda}$, $\lambda$ = 0.5)~\cite{Fan2017Smeasure},
and maximum E-measure ($\mathcal{E}_{\xi}$)~\cite{Fan2018Emeasure}.

\subsection{Comparison with State-of-the-art Methods}
\noindent\textbf{Comparison Methods.} 
We compare the proposed CMWNet with 6 state-of-the-art traditional methods, which are
LBE~\cite{Feng2016LBE}, DCMC~\cite{Cong2016SPL},
SE~\cite{Guo2016ICME}, CDCP~\cite{Zhu2017CDCP},
MDSF~\cite{Song2017MDSF} and DTM~\cite{Cong2019DTM},
and 9 state-of-the-art CNN-based methods, which are
DF~\cite{Qu2017DF}, CTMF~\cite{Han2018CTMF},
PCF~\cite{Chen2018PCF}, AFNet~\cite{Wang2019AFNet},
MMCI~\cite{Chen2019MMCI}, TANet~\cite{Chen2019TANet},
CPFP~\cite{Zhao2019CPFP}, DMRA~\cite{LHC2019DMRA}
and D3Net~\cite{Fan2019D3Net}.
The saliency maps of all compared methods are provided
by authors or obtained by running their released codes.
Notably, we retest DMRA~\cite{LHC2019DMRA} on STEREO dataset with 1,000 images,
which results in different performances from the original DMRA paper.
\begin{table*}[t!]
  \centering
  \small
  \renewcommand{\arraystretch}{1.8}
  \renewcommand{\tabcolsep}{0.8mm}
  \caption{\small
  \textbf{Quantitative results of 15 state-of-the-art methods on 7 datasets}:
  \textit{STEREO}~\cite{Niu2012STEREO},
  \textit{NJU2K}~\cite{Ju2014NJU2K},
  \textit{LFSD}~\cite{Li2014LFSD},
  \textit{DES}~\cite{Cheng2014DES},
  \textit{NLPR}~\cite{Peng2014NLPR},
  \textit{SSD}~\cite{SSD2017},
  and \textit{SIP}~\cite{Fan2019D3Net}.
   ``-T'' indicates the results on the test set of this dataset. 
   $\uparrow$ and $\downarrow$ stand for larger and smaller is better, respectively.
    The best two results are marked in \textcolor{red}{\textbf{red}} and \textcolor{blue}{\textbf{blue}}.
    The \textit{corner note} of each method is the publication year.
  }\label{tab:ModelScore}
  \resizebox{1\textwidth}{!}{
\begin{tabular}{l|cccc|cccc|cccc|cccc|cccc|cccc|cccc}
\midrule[1pt]    
 \multirow{2}{*}{\normalsize{Models}} 
 & \multicolumn{4}{c|}{STEREO~\cite{Niu2012STEREO}} 
 & \multicolumn{4}{c|}{NJU2K-T~\cite{Ju2014NJU2K}} 
 & \multicolumn{4}{c|}{LFSD~\cite{Li2014LFSD}} 
 & \multicolumn{4}{c|}{DES~\cite{Cheng2014DES}} 
 & \multicolumn{4}{c|}{NLPR-T~\cite{Peng2014NLPR}}
 & \multicolumn{4}{c|}{SSD~\cite{SSD2017}} 
 & \multicolumn{4}{c}{SIP~\cite{Fan2019D3Net}} \\
 \cmidrule(l){2-5} \cmidrule(l){6-9} \cmidrule(l){10-13} \cmidrule(l){14-17} \cmidrule(l){18-21} \cmidrule(l){22-25} \cmidrule(l){26-29}
             & $\mathcal{S}_{\lambda}\uparrow$ & $\mathcal{F}_{\beta}\uparrow$ &$\mathcal{E}_{\xi}\uparrow$ & $ \mathcal{M}\downarrow$
   	     & $\mathcal{S}_{\lambda}\uparrow$ & $\mathcal{F}_{\beta}\uparrow$ &$\mathcal{E}_{\xi}\uparrow$ & $ \mathcal{M}\downarrow$
	     & $\mathcal{S}_{\lambda}\uparrow$ & $\mathcal{F}_{\beta}\uparrow$ &$\mathcal{E}_{\xi}\uparrow$ & $ \mathcal{M}\downarrow$
	     & $\mathcal{S}_{\lambda}\uparrow$ & $\mathcal{F}_{\beta}\uparrow$ &$\mathcal{E}_{\xi}\uparrow$ & $ \mathcal{M}\downarrow$
	     & $\mathcal{S}_{\lambda}\uparrow$ & $\mathcal{F}_{\beta}\uparrow$ &$\mathcal{E}_{\xi}\uparrow$ & $ \mathcal{M}\downarrow$
	     & $\mathcal{S}_{\lambda}\uparrow$ & $\mathcal{F}_{\beta}\uparrow$ &$\mathcal{E}_{\xi}\uparrow$ & $ \mathcal{M}\downarrow$
	     & $\mathcal{S}_{\lambda}\uparrow$ & $\mathcal{F}_{\beta}\uparrow$ &$\mathcal{E}_{\xi}\uparrow$ & $ \mathcal{M}\downarrow$ \\
\midrule[1pt]
\textbf{LBE$_{16}$}~\cite{Feng2016LBE} &.660 & .633 & .787 & .250 & .695 & .748 & .803 & .153 & .736 & .726 & .804 & .208 & .703 & .788 & .890 & .208 
								 & .762 & .745 & .855 & .081 & .621 & .619 & .736 & .278 & .727 & .751 & .853 & .200 \\
\textbf{DCMC$_{16}$}~\cite{Cong2016SPL} &.731 & .740 & .819 & .148 & .686 & .715 & .799 & .172 & .753 & .817 & .856 & .155 & .707 & .666 & .773 & .111 
								 & .724 & .648 & .793 & .117 & .704 & .711 & .786 & .169 & .683 & .618 & .743 & .186 \\
\textbf{SE$_{16}$}~\cite{Guo2016ICME} &.708 & .755 & .846 & .143 & .664 & .748 & .813 & .169 & .698 & .791 & .840 & .167 & .741 & .741 & .856 & .090 
								 & .756 & .713 & .847 & .091 & .675 & .710 & .800 & .165 & .628 & .661 & .771 & .164 \\
\textbf{CDCP$_{17}$}~\cite{Zhu2017CDCP} &.713 & .664 & .786 & .149 & .669 & .621 & .741 & .180 & .717 & .703 & .786 & .167& .709 & .631 & .811 & .115 
								 & .727 & .645 & .820 & .112 & .603 & .535 & .700 & .214 & .595 & .505 & .721 & .224 \\
\textbf{MDSF$_{17}$}~\cite{Song2017MDSF} &.728 & .719 & .809 & .176 & .748 & .775 & .838 & .157 & .700 & .783 & .826 & .190 & .741 & .746 & .851 & .122 
								 & .805 & .793 & .885 & .095 & .673 & .703 & .779 & .192 & .717 & .698 & .798 & .167 \\
\textbf{DTM$_{19}$}~\cite{Cong2019DTM} &.747 & .743 & .837 & .168 & .706 & .716 & .799 & .190 & .783 & .825 & .853 & .160 & .752 & .697 & .858 & .123 
								 & .733 & .677 & .833 & .145 & .677 & .651 & .773 & .199 & .690 & .659 & .778 & .203 \\
\hline
\textbf{DF$_{17}$}~\cite{Qu2017DF} &.757 & .757 & .847 & .141 & .763 & .804 & .864 & .141 & .791 & .817 & .865 & .138 & .752 & .766 & .870 & .093 
								 & .802 & .778 & .880 & .085 & .747 & .735 & .828 & .142 & .653 & .657 & .759 & .185 \\
\textbf{CTMF$_{18}$}~\cite{Han2018CTMF} &.848 & .831 & .912 & .086 & .849 & .845 & .913 & .085 & .796 & .791 & .865 & .119 & .863 & .844 & .932 & .055 
								 & .860 & .825 & .929 & .056 & .776 & .729 & .865 & .099 & .716 & .694 & .829 & .139 \\
\textbf{PCF$_{18}$}~\cite{Chen2018PCF} &.875 & .860 & .925 & .064 & .877 & .872 & .924 & .059 & .794 & .779 & .835 & .112 & .842 & .804 & .893 & .049 
								 & .874 & .841 & .925 & .044 & .841 & .807 & .894 & .062 & .842 & .838 & .901 & .071 \\
\textbf{AFNet$_{19}$}~\cite{Wang2019AFNet} &.825 & .823 & .887 & .075 & .772 & .775 & .853 & .100 & .738 & .744 & .815 & .133 & .770 & .728 & .881 & .068 
								 & .799 & .771 & .879 & .058 & .714 & .687 & .807 & .118 & .720 & .712 & .819 & .118 \\
\textbf{MMCI$_{19}$}~\cite{Chen2019MMCI} &.873 & .863 & .927 & .068 & .858 & .852 & .915 & .079 & .787 & .771 & .839 & .132 
								 & .848 & .822 & .928 & .065 & .856 & .815 & .913 & .059 & .813 & .781 & .882 & .082 & .833 & .818 & .897 & .086 \\
\textbf{TANet$_{19}$}~\cite{Chen2019TANet} &.871 & .861 & .923 & .060 & .878 & .874 & .925 & .060 & .801 & .796 & .847 & .111 & .858 & .827 & .910 & .046 
								 & .886 & .863 & .941  & .041 & .839 & .810 & .897 & .063 & .835 & .830 & .895 & .075 \\
\textbf{CPFP$_{19}$}~\cite{Zhao2019CPFP} & .879 & .874 & .925 & \textcolor{blue}{\textbf{.051}} & .878 & .877 & .923 & .053 & .828 & .826 & .872 & .088 & .872 & .846 & .923 & .038 
								 & .888 & .867 & .932 & .036 & .807 & .766 & .852 & .082 & .850 & .851 & \textcolor{blue}{\textbf{.903}} & .064 \\
\textbf{DMRA$_{19}$}~\cite{LHC2019DMRA} & .835 & .847 & .911 & .066 & .886 & .886 & .927 & \textcolor{blue}{\textbf{.051}} 
								 & \textcolor{blue}{\textbf{.847}} & \textcolor{blue}{\textbf{.856}} & \textcolor{blue}{\textbf{.900}} & \textcolor{blue}{\textbf{.075}} 
								 & .900 & \textcolor{blue}{\textbf{.888}} & .943 & \textcolor{blue}{\textbf{.030}}  
							 	 & .899 & .879 & \textcolor{blue}{\textbf{.947}} & \textcolor{blue}{\textbf{.031}} 
								 & .857 & .844 & .906 & \textcolor{blue}{\textbf{.058}} & .806 & .821 & .875 & .085 \\
\textbf{D3Net$_{19}$}~\cite{Fan2019D3Net} & \textcolor{blue}{\textbf{.891}} & \textcolor{blue}{\textbf{.881}} & \textcolor{blue}{\textbf{.930}} & .054 
								 & \textcolor{blue}{\textbf{.895}} & \textcolor{blue}{\textbf{.889}} & \textcolor{blue}{\textbf{.932}} & \textcolor{blue}{\textbf{.051}} 
								 & .832 & .819 & .864 & .099 
								 & \textcolor{blue}{\textbf{.904}} & .885 & \textcolor{blue}{\textbf{.946}} & \textcolor{blue}{\textbf{.030}} 
								 & \textcolor{blue}{\textbf{.906}} & \textcolor{blue}{\textbf{.885}} & .946 & .034 
								 & \textcolor{blue}{\textbf{.866}} & \textcolor{blue}{\textbf{.847}} & \textcolor{blue}{\textbf{.910}} & \textcolor{blue}{\textbf{.058}} 
								 & \textcolor{blue}{\textbf{.864}} & \textcolor{blue}{\textbf{.862}} & \textcolor{blue}{\textbf{.903}} & \textcolor{blue}{\textbf{.063}} \\
\hline
\hline
\textbf{Ours} & \textcolor{red}{\textbf{.905}} & \textcolor{red}{\textbf{.901}} & \textcolor{red}{\textbf{.944}} & \textcolor{red}{\textbf{.043}}
			     & \textcolor{red}{\textbf{.903}} & \textcolor{red}{\textbf{.902}} & \textcolor{red}{\textbf{.936}} & \textcolor{red}{\textbf{.046}}
			     & \textcolor{red}{\textbf{.876}} & \textcolor{red}{\textbf{.883}} & \textcolor{red}{\textbf{.912}} & \textcolor{red}{\textbf{.066}}
			     & \textcolor{red}{\textbf{.934}} & \textcolor{red}{\textbf{.930}} & \textcolor{red}{\textbf{.969}} & \textcolor{red}{\textbf{.022}}
			     & \textcolor{red}{\textbf{.917}} & \textcolor{red}{\textbf{.903}} & \textcolor{red}{\textbf{.951}} & \textcolor{red}{\textbf{.029}}
			     & \textcolor{red}{\textbf{.875}} & \textcolor{red}{\textbf{.871}} & \textcolor{red}{\textbf{.930}} & \textcolor{red}{\textbf{.051}}
			     & \textcolor{red}{\textbf{.867}} & \textcolor{red}{\textbf{.874}} & \textcolor{red}{\textbf{.913}} & \textcolor{red}{\textbf{.062}} \\
\toprule[1pt]
\end{tabular}
  }
\end{table*}
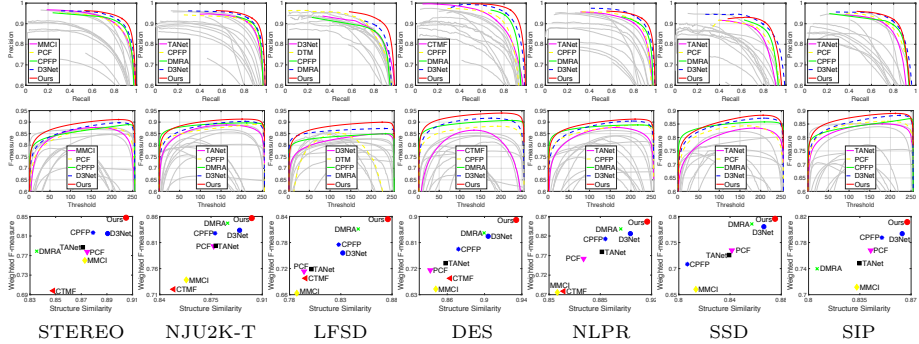
\begin{figure}[t!]
	\centering
    \small
	\begin{overpic}[width=.99\textwidth]{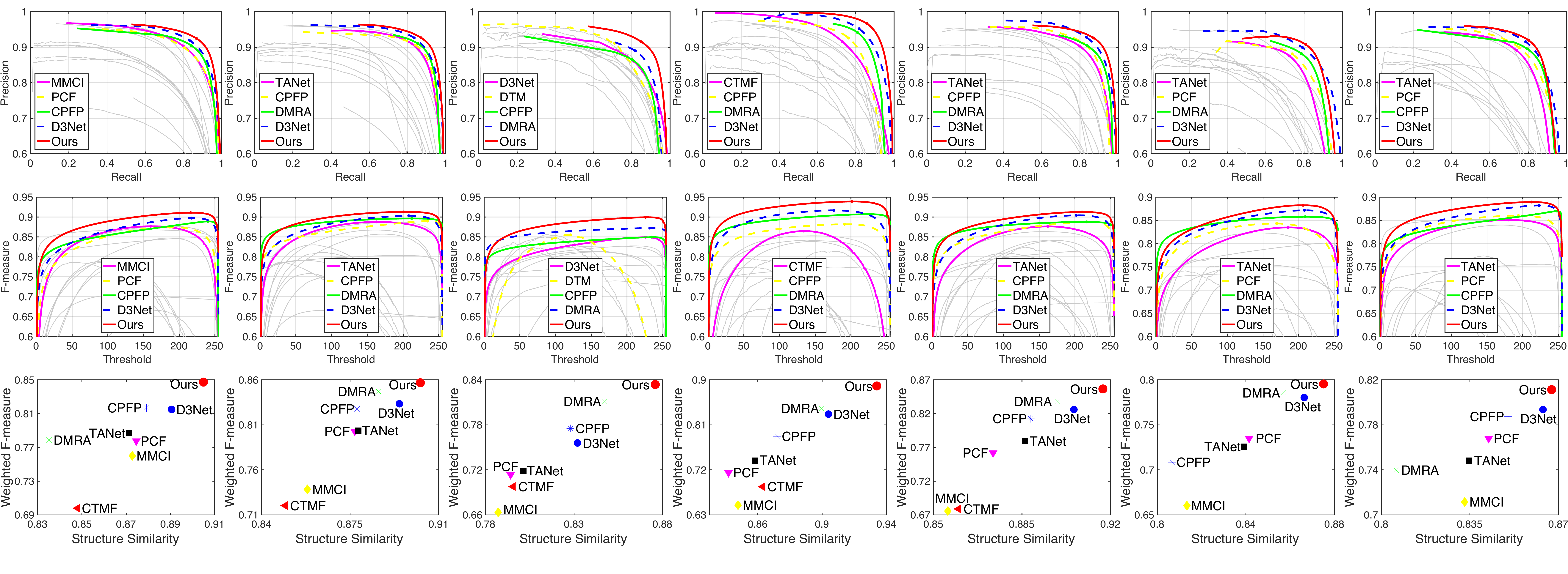}
    \put(3.3,-1){\rotatebox{0}{\scriptsize STEREO}}
    \put(17.3,-1){\rotatebox{0}{\scriptsize NJU2K-T}}
    \put(33.6,-1){\rotatebox{0}{\scriptsize LFSD}}
    \put(48.8,-1){\rotatebox{0}{\scriptsize DES}}
    \put(62.1,-1){\rotatebox{0}{\scriptsize NLPR}}
    \put(77.2,-1){\rotatebox{0}{\scriptsize SSD}}
    \put(91.8,-1){\rotatebox{0}{\scriptsize SIP}}
    \end{overpic}
	\caption{\small \textbf{Quantitative comparisons on PR curve, F-measure curve and $\mathcal{S}_{\lambda}$-$\mathcal{F}_{\beta}^w$ coordinates.}
    The top 5 methods on PR and F-measure curves are shown in color.
    For $\mathcal{S}_{\lambda}$-$\mathcal{F}_{\beta}^w$ coordinates, we only compare with several representative methods.
    }
    \label{fig:PR_Fm_wFmSm}
\end{figure}

\noindent\textbf{Quantitative Comparison.}
We evaluate our method and the other 15 state-of-the-art methods
under four quantitative metrics, including S-measure $\mathcal{S}_{\lambda}$,
max F-measure $\mathcal{F}_{\beta}$,
max E-measure $\mathcal{E}_{\xi}$ and MAE $\mathcal{M}$.
As shown in Tab.~\ref{tab:ModelScore}, our method favorably
outperforms all compared methods under these four metrics,
and the recently proposed CNN-based methods
\cite{Chen2019TANet,Zhao2019CPFP,LHC2019DMRA,Fan2019D3Net}
and our method are superior to traditional methods by a large margin.
Comparing to the second best results in Tab.~\ref{tab:ModelScore},
the performance of our method on the largest and challenging dataset
STEREO~\cite{Niu2012STEREO} is improved by 1.6\% and 2.0\%
in $\mathcal{S}_{\lambda}$ and $\mathcal{F}_{\beta}$, respectively.
The improvement on the relatively small dataset DES~\cite{Cheng2014DES}
is remarkable, with an increase of 3.0\% and 4.2\%
in $\mathcal{S}_{\lambda}$ and $\mathcal{F}_{\beta}$, respectively.
For the salient person detection, our method improves the performance
by 1.2\% in $\mathcal{F}_{\beta}$ on SIP~\cite{Fan2019D3Net}.
In addition, we also present the PR curves, F-measure curves and
$\mathcal{S}_{\lambda}$ (X-axis) -$\mathcal{F}_{\beta}^w$ (Y-axis) coordinates
in Fig.~\ref{fig:PR_Fm_wFmSm}.
The performance under these metrics is
consistent with that in Tab.~\ref{tab:ModelScore}.
The superiority of our method is more visible on
STEREO~\cite{Niu2012STEREO}, LFSD~\cite{Li2014LFSD}
and DES~\cite{Cheng2014DES}.
Both Tab.~\ref{tab:ModelScore} and Fig.~\ref{fig:PR_Fm_wFmSm}
demonstrate that our method is consistently better than all compared
methods in terms of different evaluation metrics.

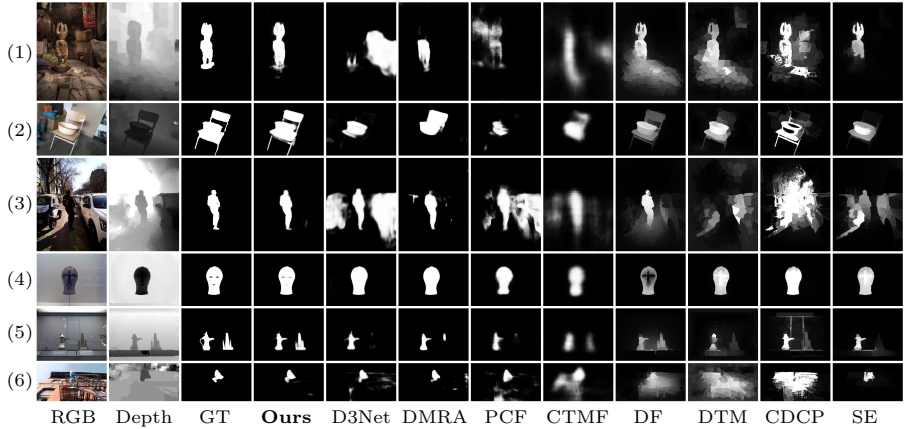
\begin{figure*}[t!]
    \centering
    \small
	\begin{overpic}[width=.96\textwidth]{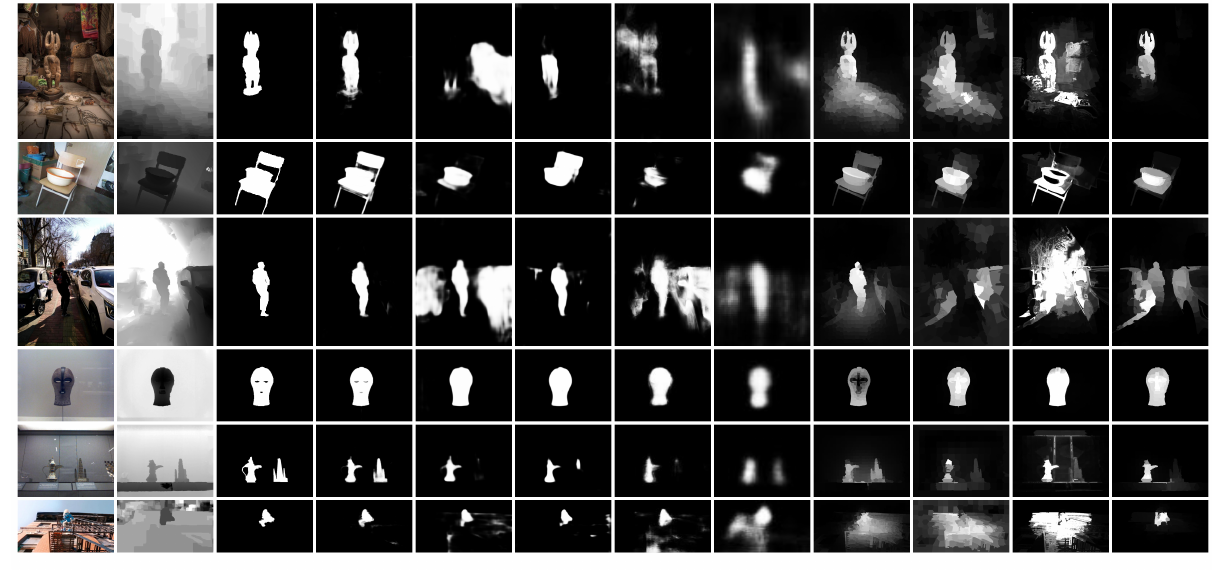}
    \put(-2,40.5){\scriptsize{(1)}}
    \put(-2,31.6){\scriptsize (2)}
    \put(-2,23.3){\scriptsize (3)}
    \put(-2,14.7){\scriptsize (4)}
    \put(-2,8.6){\scriptsize (5)}
    \put(-2,3.15){\scriptsize (6)}

    \put(3,-1){\scriptsize RGB}
    \put(10.4,-1){\scriptsize Depth}
    \put(19.9,-1){\scriptsize GT}
    \put(26.9,-1){\scriptsize \textbf{Ours}}
    \put(34.9,-1){\scriptsize D3Net}
    \put(42.9,-1){\scriptsize DMRA}
    \put(52.3,-1){\scriptsize PCF}
    \put(59.3,-1){\scriptsize CTMF}
    \put(69.2,-1){\scriptsize DF}
    \put(76.5,-1){\scriptsize DTM}
    \put(84.1,-1){\scriptsize CDCP}
    \put(94,-1){\scriptsize SE}
    \end{overpic}
	\caption{\small \textbf{Visual comparisons with eight representative methods},
	including five CNN-based methods (D3Net~\cite{Fan2019D3Net},
	DMRA~\cite{LHC2019DMRA}, PCF~\cite{Chen2018PCF},
	CTMF~\cite{Han2018CTMF}, DF~\cite{Qu2017DF}) and three traditional methods
	(DTM~\cite{Cong2019DTM}, CDCP~\cite{Zhu2017CDCP}, SE~\cite{Guo2016ICME}).
    }
    \label{fig:VisualExample}
\end{figure*}

\noindent\textbf{Visual Comparison.}
We show visual comparisons with 8 representative
methods in Fig.~\ref{fig:VisualExample}.
Each row in Fig.~\ref{fig:VisualExample} represents a
challenging scenario for SOD, including low contrast  ($1^{st}$ row),
disturbing background ($2^{nd}$ row), salient person detection ($3^{rd}$ row),
object with fine structures ($4^{th}$ row), multiple objects ($5^{th}$ row)
and small object ($6^{th}$ row).
Regardless of different scene, our method can accurately highlight
salient objects with fine details.
Notably, in the $4^{th}$ row, the mask has a fine structure
with three holes on it.
Thanks to the DW and RW operations in our method,
our method successfully highlights the mask with three holes,
while other methods fail.

\subsection{Ablation Studies}
\label{sec:AblationStudy}

We conduct detailed ablation studies of our CMWNet on
a big dataset, NJU2K~\cite{Ju2014NJU2K}, and
a small but challenging dataset, SSD~\cite{SSD2017}.
Specifically, we assess
1) the rationality of enhancing RGB features with depth features;
2) the individual contributions of CMW-L\&M and CMW-H;
3) the importance of weighting;
4) the rationality of cross-scale weighting of CMW-L\&M;
and 5) the necessity of deep scale supervision of decoder part.
We change one component at a time to evaluate individual contributions.

\noindent\textbf{Rationality of Enhancing RGB Features with Depth Features.}
In our method, we use depth features to enhance RGB features (``DeR").
To study the rationality of this enhancement manner,
we explore another baseline variant: adopting RGB features
to enhance depth features (``ReD").
From Tab.~\ref{tab:AblationStudy}, we observe that the performance on
both datasets has dropped (\eg
~$\mathcal{M}\!: 0.046\!\rightarrow\!0.056$ on NJU2K and
$0.051\!\rightarrow\!0.063$ on SSD).
This confirms that using depth features to enhance RGB features is more
reasonable than the other direction for extracting the cross-modal complementarity.

In addition, we remove the depth map input in our network to evaluate the power
of depth map (\textit{w/o} depth).
This variant is for RGB SOD.
The performance of \textit{w/o} depth drops sharply
(\eg~$\mathcal{M}\!: 0.046\!\rightarrow\!0.056$ on NJU2K).
This confirms that the way of exploring complementary distance information of depth map
in our network is effective.

\begin{table}[t!]
  \centering
  \renewcommand{\arraystretch}{1.4}
  \renewcommand{\tabcolsep}{2mm}
  \caption{\small
  \textbf{Ablation studies on \textit{NJU2K}~\cite{Ju2014NJU2K} and \textit{SSD}~\cite{SSD2017}.}
    The best result of each column is \textbf{bold}.
    Details are introduced in \S~\ref{sec:AblationStudy}.    
  }\label{tab:AblationStudy}
  \resizebox{0.65\textwidth}{!}{
\begin{tabular}{c||cccc|cccc}
\bottomrule[1pt]
 \multirow{2}{*}{Models} 
 & \multicolumn{4}{c|}{NJU2K-T~\cite{Ju2014NJU2K}} 
 & \multicolumn{4}{c}{SSD~\cite{SSD2017}} \\
\cline{2-9}
    & $\mathcal{S}_{\lambda}\uparrow$ & $\mathcal{F}_{\beta}\uparrow$ &$\mathcal{E}_{\xi}\uparrow$ & $ \mathcal{M}\downarrow$
    & $\mathcal{S}_{\lambda}\uparrow$ & $\mathcal{F}_{\beta}\uparrow$ &$\mathcal{E}_{\xi}\uparrow$ & $ \mathcal{M}\downarrow$ \\ 
\hline
\hline
\textbf{Ours} (DeR) & ${\textbf{.903}}$ & ${\textbf{.902}}$ & ${\textbf{.936}}$ & ${\textbf{.046}}$ 
			     & ${\textbf{.875}}$ & ${\textbf{.871}}$ & ${\textbf{.930}}$ & ${\textbf{.051}}$  \\
\hline
\hline
{ReD} & $.889$ & $.887$ & $.927$ & $.056$ 
		        & $.864$ & $.850$ & $.909$ & $.063$  \\
\textit{w/o} depth (\textit{w/o} DW) & $.886$ & $.886$ & $.924$ & $.056$
		        & $.855$ & $.842$ & $.915$ & $.064$  \\
\hline
\textit{w/o} CMW-L\&M & $.891$ & $.886$ & $.932$ & $.053$
		        & $.849$ & $.839$ & $.909$ & $.066$  \\
\textit{w/o} CMW-H & $.896$ & $.894$ & $.929$ & $.051$
		        & $.853$ & $.845$ & $.908$ & $.063$  \\
\hline
\textit{w/o} RW & $.901$ & $.899$ & $.933$ & $.046$
		        & $.868$ & $.861$ & $.919$ & $.056$  \\
\textit{w/o} Wei & $.900$ & $.898$ & $.933$ & $.048$ 
		        & $.858$ & $.839$ & $.899$ & $.061$  \\
		        
DW \textit{w/o} GF & $.900$ & $.898$ & $.933$ & $.048$ 
		        & $.868$ & $.858$ & $.923$ & $.054$  \\		        
RW \textit{w/} GF & $.901$ & $.900$ & $.934$ & $.046$
		        & $.870$ & $.867$ & $.924$ & $.052$  \\

\hline
\textit{w/o} CS & $.901$ & $.898$ & $.932$ & $.047$ 
		        & $.864$ & $.861$ & $.922$ & $.060$  \\
C2S & $.900$ & $.899$ & $.933$ & $.049$
		        & $.864$ & $.847$ & $.906$ & $.060$  \\
\hline
\textit{w/o} DS & $.898$ & $.898$ & $.933$ & $.049$ 
		        & $.866$ & $.862$ & $.923$ & $.055$  \\
\hline

\end{tabular}
}
\end{table}

\noindent\textbf{Individual Contributions of CMW-L\&M and CMW-H.}
The proposed three RGB-depth interaction modules can be divided into two types.
CMW-L and CMW-M (CMW-L\&M) are responsible for object details enhancement, and
CMW-H for object localization enhancement.
Thus, we provide two variants of our network:
removing the CMW-L\&M (\textit{w/o} CMW-L\&M) and
removing the CMW-H (\textit{w/o} CMW-H).
From Tab.~\ref{tab:AblationStudy}, we observe a significant performance degradation
(\eg~$\mathcal{S}_{\lambda}\!: 0.903\!\rightarrow\!0.891$ on NJU2K and
$0.875\!\rightarrow\!0.849$ on SSD) of \textit{w/o} CMW-L\&M.
This confirms that the proposed CMW-L\&M are momentous to our network, and they enhance
the details of salient object in low- and middle-level features clearly.
Some enhanced features in \textbf{CMW-L} are shown in the third column of
Fig.~\ref{fig:FeaturemapVisualize}.
The performance drop (\eg~$\mathcal{F}_{\beta}\!: 0.902\!\rightarrow\!0.894$ on NJU2K
and $0.871\!\rightarrow\!0.845$ on SSD)
of \textit{w/o} CMW-H means that the proposed CMW-H is also important to our network
and it enhances the salient object localization in high-level features accurately.

\noindent\textbf{Importance of Weighting.}
To study the importance of two types of weighting, we derive three variants:
removing the Depth-to-RGB weighting (\textit{w/o} DW),
removing the RGB-to-RGB weighting (\textit{w/o} RW),
and using concatenation of depth features and RGB features instead of
two types of weighting (\textit{w/o} Wei).
Specifically, \textit{w/o} DW is the same as \textit{w/o} depth, \ie,
the depth map does not participate in SOD.
The depth map is still utilized in \textit{w/o} Wei, which is not equal to \textit{w/o} (DW+RW).
It focuses on evaluating the impact of weighting mechanism.
According to the statistics in Tab.~\ref{tab:AblationStudy},
we observe the performance of these three variants is
worse than our complete CMWNet.
This demonstrates that the two types of weighting can help our CMWNet to
better highlight salient objects with effective feature enhancement.
We also provide two variants, \ie~DW \textit{w/o} GF and RW \textit{w/} GF,
to confirm the rationality of specific global filters (GF) in DW.

In addition, the visualization of features \textit{w/o} DW and
\textit{w/o} RW is shown in Fig.~\ref{fig:FeaturemapVisualize},
in which the boundaries of salient objects \textit{w/o} RW are
much clearer than \textit{w/o} DW.
This demonstrates that the depth map does assist the RGB-D SOD,
and the enhancement effect of DW is more effective than RW.

\noindent\textbf{Rationality of Cross-Scale Weighting in CMW-L\&M.}
To study the rationality of cross-scale weighting in CMW-L\&M,
we modify the cross-scale DW to the same-scale
manner (\textit{w/o} CS), which is the same as CMW-H.
The double enhanced features of each scale in CMW-L\&M
are concatenated for inference.
By comparing \textit{w/o} CS and Ours in Tab.~\ref{tab:AblationStudy},
we find that the performance of \textit{w/o} CS decreases on both
NJU2K and SSD.
This demonstrates that the DW of CMW-L\&M in the
cross-scale manner is rational, and this manner can enhance
interactions between different scales to further
boost performance.

Besides, to study the rationality of performing cross-scale weighting
between adjacent CNN blocks, we provide a variant which performs cross-scale
weighting between nonadjacent CNN blocks,
\ie, R-E$^{(1)}$ is enhanced by D-E$^{(3)}$,
R-E$^{(2)}$ is enhanced by D-E$^{(4)}$,
R-E$^{(3)}$ is enhanced by D-E$^{(1)}$ and
R-E$^{(4)}$ is enhanced by D-E$^{(2)}$ (C2S).
As the results presented in Tab.~\ref{tab:AblationStudy}, we observe that the
results of C2S are worse than Ours.
The reason behind this is that the weighting performed across two scales (\ie~C2S) 
may lose the continuity of features, causing the depth response maps to fail to
highlight the salient objects of RGB features.
In contrast, the weighting performed between two adjacent CNN blocks (\ie~Ours)
can capture the continuity of features and precisely increase cross-scale interactions.

\noindent\textbf{Necessity of Deep Scale Supervision in Decoder.}
To study the necessity of deep scale supervision, we provide a baseline
with only one supervision of the final prediction $S^{(1)}$ (\textit{w/o} DS).
As shown in Tab.~\ref{tab:AblationStudy}, we observe that network training
with additional supervision is better than the single supervision.
This verifies that the multiple scale supervision during network training can
improve the testing performance.
Besides, the intermediate predictions $S^{(4)}$, $S^{(3)}$ and $S^{(2)}$
are also shown in Fig.~\ref{fig:Framework}.
We can observe that the refinement process of
prediction from coarse ($S^{(4)}$) to fine ($S^{(1)}$) benefits from the deep scale
supervision in decoder part, which visually confirms the necessity of deep scale supervision.

\section{Conclusion}

In this paper, we propose a novel Cross-Modal Weighting Network (CMWNet) for RGB-D SOD. In particular, three novel cross-modal cross-scale weighting modules (CWM-L, CMW-M and CMW-H) are designed to encourage feature interactions for improving SOD performance. 
Based on these improvements, a three-level decoder progressively refines salient objects.
Extensive experiments are conducted to validate our CMWNet, which achieves the best performance on seven public RGB-D SOD benchmarks in comparison with 15 state-of-the-arts.

\noindent\textbf{Acknowledgments.}
This work was supported by the National Natural Science Foundation of China under Grant 61771301.
Linwei Ye and Yang Wang were supported by NSERC.

%
%
\bibliographystyle{splncs04}
\bibliography{RGBDref}

\end{document}